# Multimodal Perception System for Real Open Environment

Yuyang Sha

Faculty of Applied Sciences, Macao Polytechnic University,

Macau SAR, 999078, China

**ABSTRACT**

This paper presents a novel multimodal perception system for a real open environment. The proposed system includes an embedded computation platform, cameras, ultrasonic sensors, GPS, and IMU devices. Unlike the traditional frameworks, our system integrates multiple sensors with advanced computer vision algorithms to help users walk outside reliably. The system can efficiently complete various tasks, including navigating to specific locations, passing through obstacle regions, and crossing intersections. Specifically, we also use ultrasonic sensors and depth cameras to enhance obstacle avoidance performance. The path planning module is designed to find the locally optimal route based on various feedback and the user's current state. To evaluate the performance of the proposed system, we design several experiments under different scenarios. The results show that the system can help users walk efficiently and independently in complex situations.

## 1. INTRODUCTION

In recent years, rapid advancements in artificial intelligence have led to groundbreaking developments in various cutting-edge technologies. These innovative technologies have become prevalent in our everyday lives, impacting areas such as autonomous driving[1], face perception[2], [3], and medical data analysis[4]. In recent years, there has been a remarkable surge in the development and application of advanced artificial intelligence technologies, such as ChatGPT . These technologies [5], [6], [7] have exhibited exceptional transfer learning capabilities across various language-related tasks. Consequently, there has been a noticeable shift in research focus toward multimodal models, indicating a growing interest in models capable of effectively processing and interpreting various forms of data. Different from the focus of researchers, major technology companies have started incorporating multimodal technologies into their product designs. For example, Apple has invested heavily in Metaverse devices and successfully developed Apple Vision Pro, which can make user interact with their favorite apps, capture and relive memories, and enjoy stunning TV shows and movies. The widespread adoption of numerous VR and MR devices has notably influenced gaming and social networking [8]. Multimodal perception [9], which integrates information from multiple sensory modalities such as vision, radar, and LiDAR, plays a crucial role in autonomous driving systems . It is a fundamental technology that enables vehicles to perceive and understand their surroundings, enhancing safety and decision-making capabilities. However, few people pay attention to pedestrian-centric multimodal perception technology, which can help us understand the surrounding environment and lay a solid foundation for the further development of metaverse technology. Furthermore, pedestrian-based multimodal perception technology holds substantial potential in addressing practical needs such as enhancing medical assistance and enabling navigation for the visually impaired.

In this paper, we present a novel multimodal perception that can solve the problems of environmental perception, obstacle avoidance, and navigating for users in real open environments. With the help of our system, users can walk independently in unfamiliar and complex places like the sighted ones. Similar to human walking, our system consists of four parts: information collection, environmental perception, path planning, and human-robot interaction. We employ various sensors for collecting environmental information, such as RGB-D camera, ultrasonic sensors, global positioning system (GPS), inertial measurement unit (IMU), etc. The environmental perception module is responsible for analyzing multimodality information collected by sensors. Vision-based algorithms are widely used in the environmental perception module. For example, the lightweight object detection model is used to find pedestrians, vehicles, traffic signals, etc. The high-efficiency segmentation method is applied to find blind tracks, sidewalks, and roadways. We introduce a cross-domain depth map completion algorithm into our system to obtain a more accurate dense depth map. In addition, the proposed method could also get the current speed and direction of the user by feedbacks from the GPS and IMU. According to the results provided by the information processing modules, the path planning carefully designs A-star [10] search algorithms to find the locally optimal route for the user in real-world open environments. Then, the human-robot interaction module guides the user to the target location via audio instructions. We design several tasks for evaluating the performance of our proposed system, including navigating to specific locations, passing through obstacle regions, and crossing intersections. The results demonstrate the usability and reliability of our proposed multimodal perception system for user in complex circumstances.

## 2. MULTIMODAL PERCEPTION SYSTEM

### 2.1 Overview

In this section, we introduce our multimodal perception system in detail. The system architecture is shown in Figure 1. It includes an embedded computer, battery, GPS module, wireless headphones, two cameras, and five ultrasonic sensors. The embedded computer is leveraged to deploy algorithms; cameras and ultrasonic sensors are used for environmental perception. Similarly, the headphones are applied for human-robot interaction.

In order to help users better understand their surroundings, we conduct thorough research on various existing navigation solutions. The proposed multimodal system includes three parts: environmental perception, path planning, and human-robot interaction. The environmental perception module plays an essential role in the proposed system. It analyzes the collected information about the surrounding environment and uses it to finish several tasks, such as obstacle avoidance and reaching a specific location. Besides, the environmental perception module can also obtain users' speed and angle based on the IMU and GPS feedback. The path planning module selects the optimal path from multiple ones. Next, we detail the proposed system from the perspectives of hardware and software, respectively.

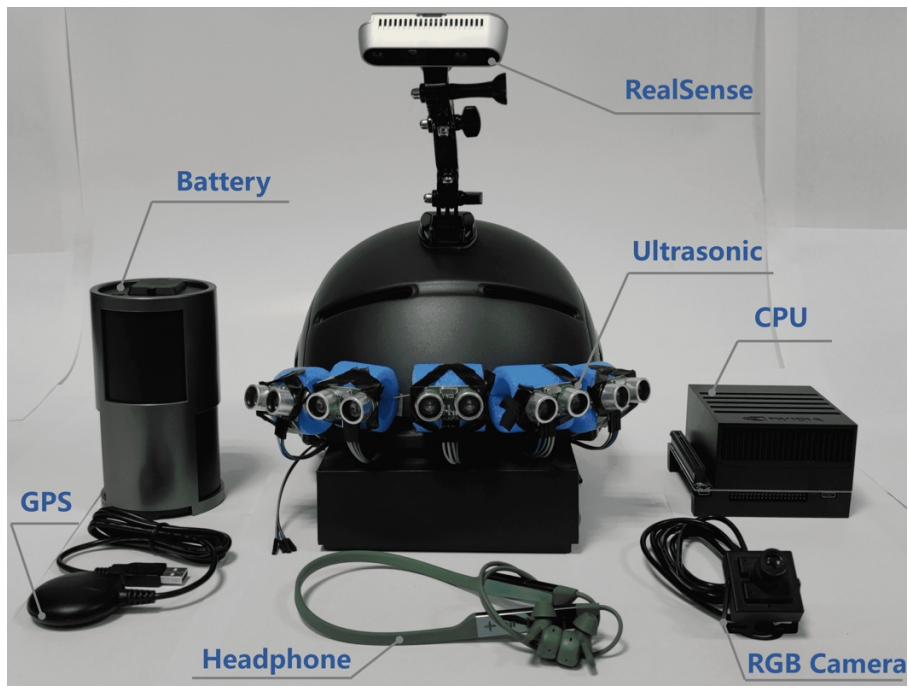

Figure 1. Details of the proposed multimodal navigation system, which includes a battery, a GPS module, a RealSense camera, five ultrasonic sensors, a headphone, an embedded computer, and an RGB camera.

### 2.2 Hardware Setup

The hardware in our system is used for information collection and deploying algorithms. We use various types of sensors to enhance the information collection capabilities of our system. The chosen cameras include two different types. One is Intel RealSense D456, which can provide RGB images and depth maps from the user perspective. The other is a high-resolution RGB-IR camera that offers 30 frames per second, and the images are in the resolution of 1920 × 1080.

The HD camera captures the images around the user's knee. The RealSense camera contains an IMU that can report the user's speed, angle rate, and body orientation. In addition, the system also integrates five ultrasonic sensors to measure the distance to obstacles in the surrounding environments. In addition, we utilize GPS and the Google Maps Platform API to assist users in completing complex tasks in open environments.

All of our system's algorithms run on an embedded computer. Generally, methods based on deep neural networks need to consume a mass of resources, but the navigation system attempts to pursue a longer battery life through low power consumption. The Nvidia Jetson Xavier NX achieves a better speed-efficiency trade-off. Therefore, we choose it as the embedding computing platform. A 100W battery powers the navigation system and can run continuously for over 5 hours.

## 2.3 Software Function

With the development of artificial intelligence, deep learning has been widely utilized in computer vision [11], generative tasks [12], and medical data analysis [13]. Deep learning methods generally outperform traditional approaches in terms of accuracy and speed [14]. Therefore, we introduce methods based on deep neural networks for environmental perception and the route search algorithm for path planning. The environmental perception module tackles tasks such as walkable region recognition, object detection, and obstacle avoidance. The path planning module uses the A-Star searching algorithm to select the locally optimal route for users.

### 2.3.1 Walkable Region Recognition

The proposed multimodal navigation system for the natural open environment attempts to help users walk independently and safely in complex environments. As a result, the system must be able to identify walkable regions such as sidewalks, roadways, and zebra crossings. Users should select appropriate, safe, and obstacle-free activity areas when moving in outdoor or indoor environments.

Since most public image segmentation datasets are used for self-driving cars, they cannot be applied directly to our task. Therefore, we collect about 500 minutes of videos containing 100 scenes such as the street, park, intersection, etc. Then, we select 5,000 images from these videos and annotate segmentation masks for walkable areas like sidewalks and roadways. Sample masks are shown in Figure 2.

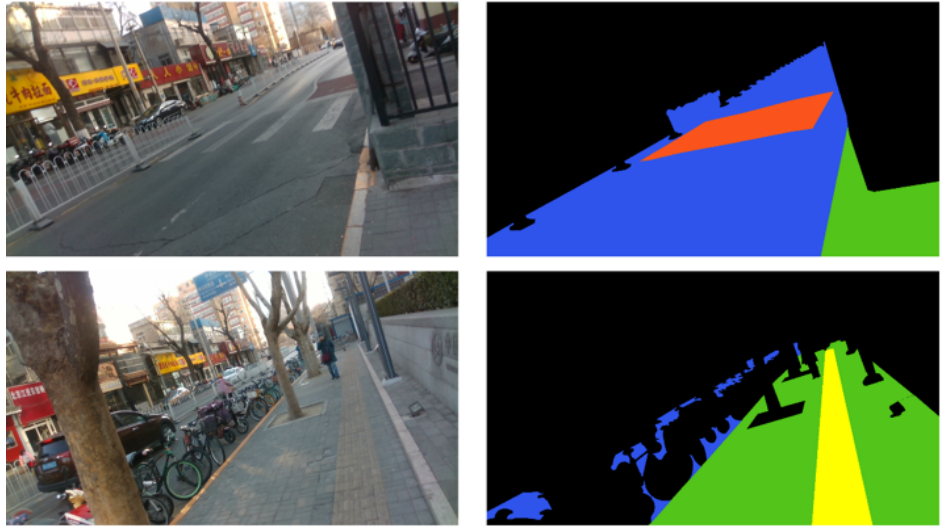

Figure 2. The annotated samples for walkable regions recognition. The original image is on the left, and the marked result is on the right.

As we all know, walkable regions recognition is highly related to image segmentation. In order to accomplish this task, we introduce a lightweight image segmentation model to our system, which could consume less power but have low latency with high accuracy. The details of proposed segmentation model are shown in Figure 3. The proposed lightweight image segmentation model is designed based on multi-scale features. It consists of three main parts: a stem block, a dual-branch feature processing module, and a multi-scale feature fusion module. In particular, the stem block is mainly responsible for quickly down sampling the image. The dual-branch feature processing module includes the detail branch, semantic branch, and feature mapping module. It can extract more representative features from the input information. The multi-scale feature fusion module discovers feature correlations to improve the robustness and stability of the segmentation algorithm.

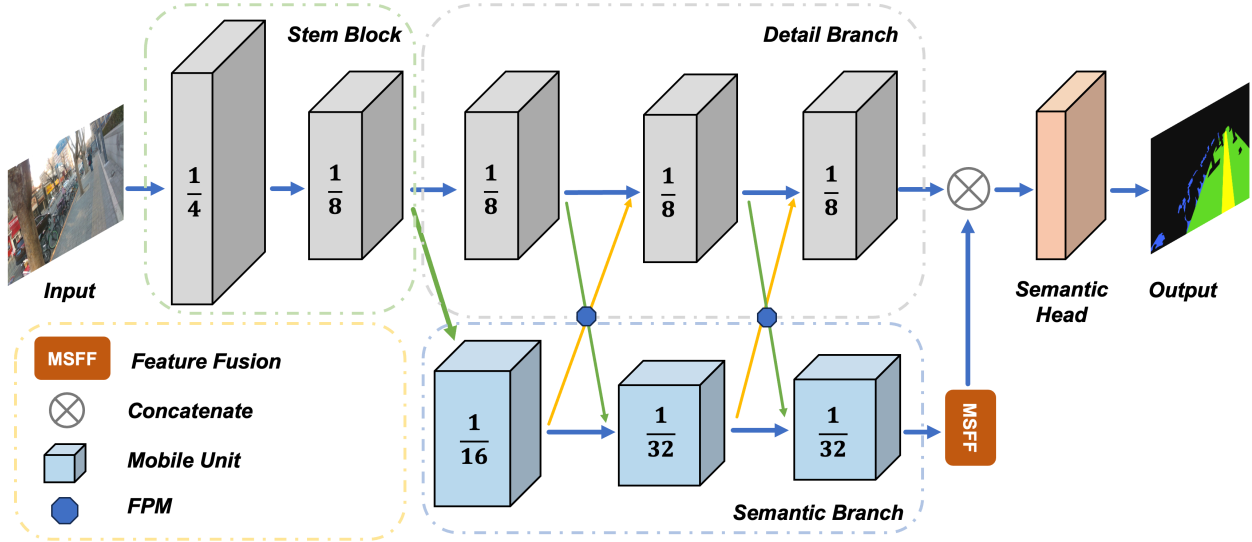

Figure 3. Overview of the proposed lightweight image segmentation model. The proposed model comprises three parts, including stem block, dual-branch feature processing module, and multi-scale feature fusion module.

We split our proposed dataset into two parts: training subset and test subset. The segmentation model is built by PyTorch with CUDA 11.6 and trained on one Nvidia A100 GPU. We train the segmentation model in two steps. In the first step, we feed the smaller image samples ($256 \times 256$) for training, and the total epochs are 100. Adopting SGD as the optimizer, its initial learning rate is 0.005. In the second step, we increase the image resolution ($512 \times 512$) and then finetune the segmentation model from the first stage for 100 epochs and using SGD with an initial learning rate of 0.001. For both these stages, we use a cosine annealing learning rate schedule for model training. Specifically, we evaluate the accuracy in terms of mean Intersection over Union (mIOU) on the test subset and compare it with other lightweight methods. Although the complex segmentation model can achieve higher accuracy, it runs slowly and cannot be used in our systems. The results are shown in Table I. We can see from the results that our model achieves a competitive performance to state-of-the-art methods while with a small computational cost. Under similar computational constraints, our model outperformance lightweight methods such as MobileSeg [15] and ESPNet-V2 [16] with a significant margin. Moreover, the accuracy of our method can meet the requirements. Figure. 4 shows some visualized segmentation results. In order to ensure safety, the user should walk slowly (about $0.4\ m/s$) in the open environment. Therefore, the segmentation model does not need to analyze each frame from the sensors. In our setting, the system conducts segmentation results every half a second, which can save resources and extend battery life.

Table 1. Comparison with lightweight segmentation methods on our dataset. We report the results in terms of mIOU(%), model parameters (M), and computational costs (FLOPs).

| Method | Year | Backbone | mIoU | Model Parameters | Computational Costs |
|---|---|---|---|---|---|
| ENet [17] | 2016 | - | 71.3 | 0.6 | 15.0 |
| ICNet [18] | 2018 | PSPNet | 82.2 | 26.5 | 14.6 |
| DeepLabV3+ [19] | 2018 | ResNet-101 | 97.6 | 62.7 | 306.5 |
| ESPNetV2 [16] | 2019 | ESPNet | 74.2 | 0.5 | 6.9 |
| BiSeNetV2 [20] | 2021 | - | 94.0 | 47.1 | 59.2 |
| MobileSeg [15] | 2023 | StrideFormer | 89.2 | 5.7 | 4.6 |
| Our Method | - | - | 94.9 | 36.5 | 140.6 |

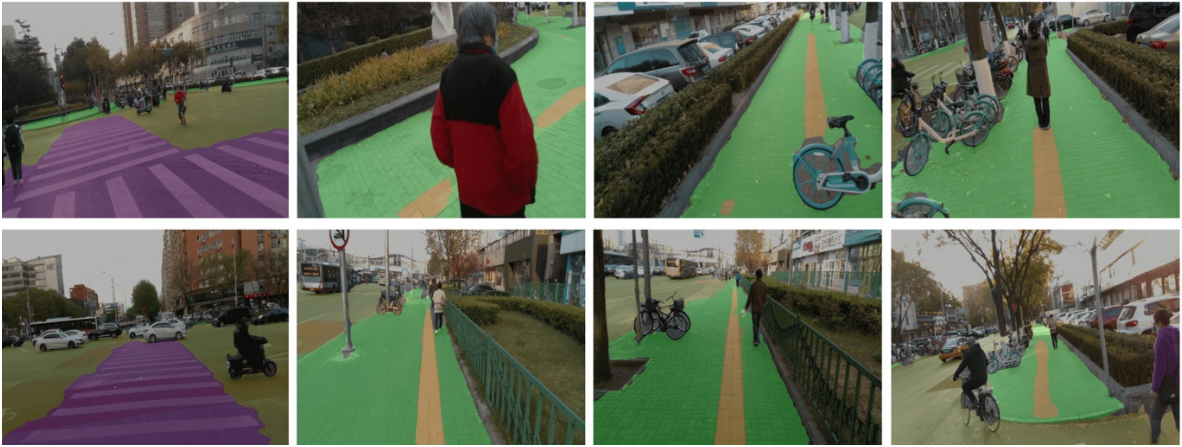

Figure 4. Visualization results of our segmentation approaches.

### 2.3.2 2D Object Detection

When the user walks in real-world open environments, the surrounding situation changes all the time. The objective of our system is to help the users to walk independently and safely in an unfamiliar environment. Therefore, the system must quickly detect the key objects that can be found in the outdoor environment. For this purpose, our system should detect seven of the most common targets in urban environments, such as people, cars, bicycles, buses, motorbikes, traffic signs, and traffic lights. To this end, we employ the object detection method based on deep neural networks to locate and recognize some specific targets in open environments. Our system adopts the YOLOv8 as the primary method, then modifies it to detect eight classes (background and seven traffic items). The training samples are selected from the COCO [21] dataset in terms of eight classes. Furthermore, we use them as the training and testing dataset. The object detection model is built by PyTorch and trained on one Nvidia A100 GPU.

The proposed multimodal navigation system has a higher requirement for the inference speed of the object detection algorithm. Therefore, we deploy our method to the embedded computational device, then test the inference speed. The results show that our approach could process 46 frames per second. Besides, the object detection algorithm is also applied for obstacle avoidance and path planning. The detection results are shown in Figure 5.

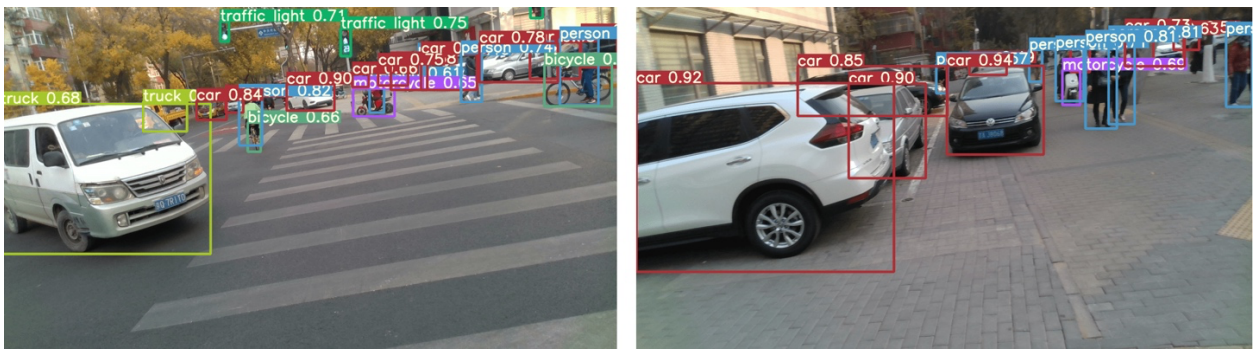

Figure 5. Visualization results of 2D object detection.

### 2.3.3 Obstacle Avoidance

Obstacle avoidance occurs when a person walks on a sidewalk and encounters obstacles, such as bicycles, cars, fire hydrants, etc. Now, some navigation systems for user solve it with ultrasonic sensors. However, it is difficult for ultrasonic sensors to locate the position of the obstacle accurately. Therefore, obstacle avoidance methods based on ultrasonic sensors are difficult to use in complex environments. Computer vision-based methods can obtain the accurate location of the target, but they are not sensitive to distance and prone to errors. In order to solve the navigation task of obstacle avoidance, our system uses multiple sensors to collect information and make full use of vision-based algorithms to enhance performance. Our system could get the depth information of the surrounding environments by the RealSense camera. For common objects such as people and cars, we use the 2D object detection method to locate the object's position in the image and then return the corresponding distance information according to the depth map. However, the performance of these methods may be degenerated severely under the outdoor environments. Therefore, we resort to leverage depth information together with five ultrasonic sensors to measure the distance between the user and surrounding targets to avoid collisions.

Since the influence of environmental factors such as lighting, some areas of the depth map might be invalid, which seriously affects the performance of the obstacle avoidance algorithm. To handle this problem, we introduce a depth completion model based on transfer learning [22], which could generate a more accurate dense depth map from a sparse depth map. We use depth-wise separable convolution instead of standard one in the backbone, thereby reducing the model size and computational costs. We train the proposed method on the KITTI depth dataset [23], then transfer the knowledge to our scene. Some results are visualized in Figure 6.

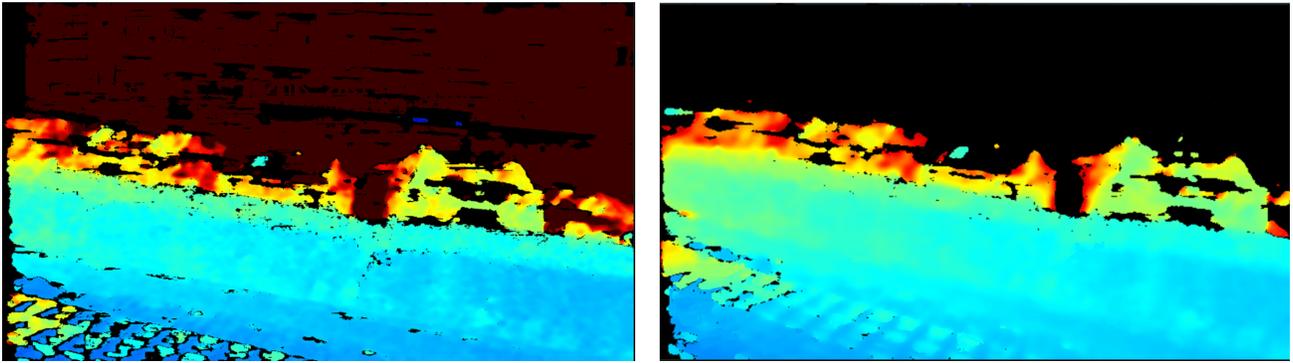

Figure 6.  The comparison of the original depth map left and completed one right.

### 2.3.4 Human-Robot Interaction

In our implementation, we choose audio for human-robot interaction. To reduce the difficulty of using our proposed system, we only employ some simple instructions generated by SqueezeWave [24] for users, such as go straight, wait, turn left, etc. Then the audio message is sent to the user via wireless headphones.

## 3.  USER STUDIES

In order to evaluate the performance of the proposed system in various scenarios for normal user, we designed three user studies, including approaching a specific location, passing through an obstacle region, and crossing an intersection. The extensive experiments are conducted on seven participants. All of the participants have no prior knowledge to this system. Participants need to study the system for about three hours. Each user is equipped with a white cane to ensure safety. The $P_i$ denotes the *i-th* participant using our system for pathing. Specifically, to ensure the safety of participants, all user studies were conducted in a controlled environment where obstacles were carefully managed to ensure the safety of users. Moreover, we also arrange staffs to protect the safety of each user.

### 3.1 Navigating to Specific Locations

To test the system's performance in outdoor scenes, we require three participants $P_3$, $P_3$, and $P_5$ to walk independently from one location to another, and the distance between the two positions is about 80 *m*. During the experiment, users need to experience various complex scenarios, such as passing through the intersection, turning left, finding the optimal route, etc. Besides, they also need to avoid pedestrians and bicycles in walkable regions.

The experimental route is shown in Figure 7-a. The results show that participants should go straight along the tracks, pass the intersection, and continue straight to reach the destination. Each participant needs to perform five times. We record the number of successfully reaching the destination, trajectories, and speed. In order to complete this task, the participant only using the white cane must get additional help, such as assisting in passing the intersection. The comparison results are shown in Table 2.

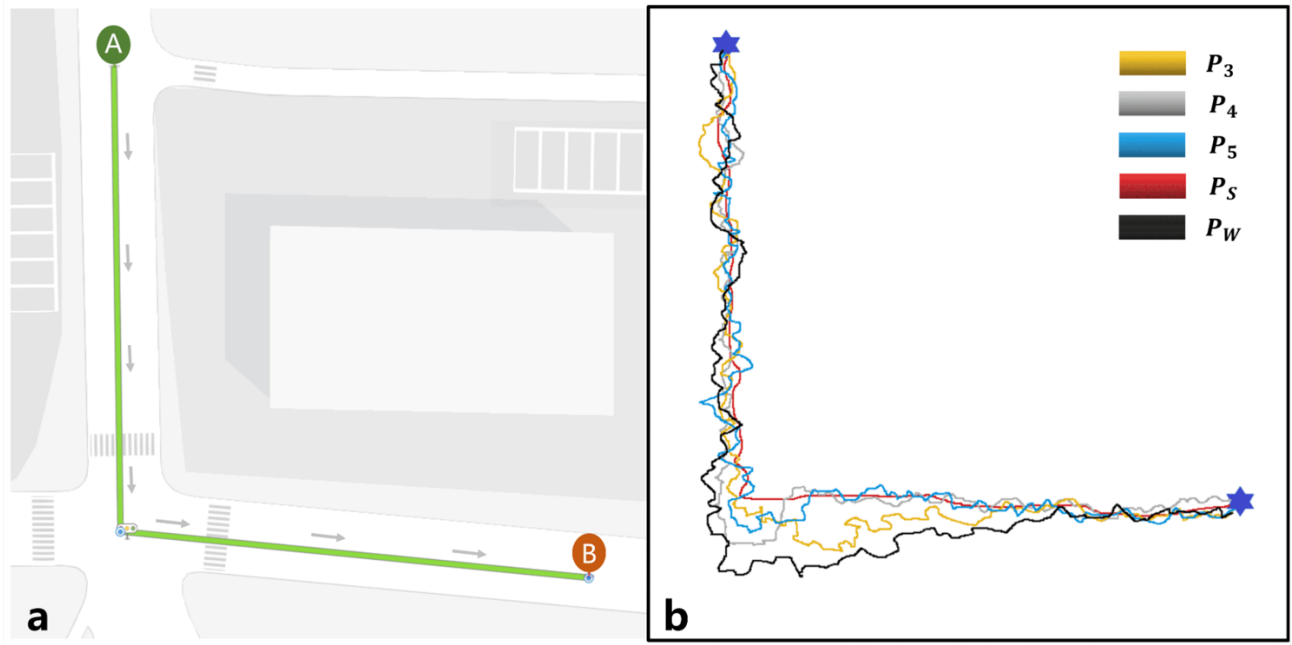

Figure 7. The comparison of the original depth map (Figure 7-a) and completed one (Figure 7-b).

It can be seen that compared with the user who is only equipped with a white cane, participant using our system have a higher success rate to reach the destination. Moreover, participants using our proposed system can walk faster. The walking trajectories are visualized in Figure. 7-b. Another participant walks in a zigzag line to adjust its orientation. However, the trajectories of participants with our system are smoother. In particular, the trajectory of $P_5$ is even close to the normal user, which demonstrates the effectiveness of our system.

Table 2. User performance in the task of navigating to specific locations.

| User | $P_3$ | $P_4$ | $P_5$ | $P_w$ | $P_s$ |
| --- | --- | --- | --- | --- | --- |
| Success Rate | 0.4 | 0.8 | 1.0 | 0.4 | 1.0 |
| Avg. Speed ($m/s$) | 0.14 | 0.21 | 0.27 | 0.08 | 0.75 |

## 3.2 Passing Through Obstacle Regions

Users should walk forward along tracks until the system finds obstacles in the visual field. Then, our system would guide the participant in avoiding obstacles. In this case, we set a specific region with six obstacles, such as bicycles and people. All of the obstacles are static. We require seven participants to pass through that region and avoid collisions. As a comparison, we ask a participant to use the white cane to complete the same task. If the user is able to pass through the obstacle region, it is regarded as a successful case. We record the time it takes for each person to complete the task and the number of collisions. Our system instructs users to avoid contact with obstacles when they pass through the obstacle region. Any collision needs to be recorded, including the body or cane contact with obstacles. Every participant should pass through the obstacle region ten times. The results in Table 3 show that our proposed system can effectively help users avoid obstacles and significantly reduce the collision rate between the user and obstacles.

Table 3. User performance in the task of passing through obstacle regions.

| User | $P_1$ | $P_2$ | $P_3$ | $P_4$ | $P_5$ | $P_w$ | $P_s$ |
| --- | --- | --- | --- | --- | --- | --- | --- |
| Success Rate | 0.5 | 0.7 | 0.7 | 0.6 | 0.9 | 0.6 | 1.0 |
| Avg. Contacts | 2.4 | 3.0 | 2.2 | 3.6 | 2.0 | 6.0 | 0 |
| Avg. Time ($s$) | 183 | 192 | 173 | 181 | 166 | 228 | 31 |

### 3.3 Crossing Intersections

Crossing an intersection is a challenging task, and it has high requirements for cooperation between the various modules in the system. For instance, the system needs to detect and recognize zebra crossings and guide users to avoid pedestrians and vehicles. We evaluate this study in a specific area with a zebra crossing length of 10 $m$ and the green light time is set to 60 seconds. Our workers simulate common situations at intersections such as pedestrians, cyclists, etc. If the user can cross the intersection within 60 seconds, it is regarded as a successful case. We record the number of users successfully passing through the intersection and the elapsed time. The results are shown in Table 4. In particular, the participant who only uses the white cane cannot pass through the intersection independently. Therefore, we only compare the results with the sighted user. Every participant should pass through the obstacle region five times. It can be seen from the results that with the help of our system, participants can independently cross the intersection. Furthermore, the successful rate of $P_5$ is up to 100%, which is even consistent to the sighted user.

Table 4. User performance in the task of navigating to specific locations.

| User | $P_1$ | $P_2$ | $P_3$ | $P_4$ | $P_5$ | $P_s$ |
|---|---|---|---|---|---|---|
| Success Rate | 0.6 | 0.4 | 0.6 | 0.8 | 1.0 | 1.0 |
| Avg. Time ($s$) | 51.9 | 55.7 | 49.5 | 51.8 | 41.9 | 9.7 |

## 4. CONCLUSION

We propose a multimodal navigation system for real world environment, which equips multiple sensors with advanced computer vision algorithms to help users walk efficiently and independently in an unfamiliar outdoor or indoor environment. The novel system can complete challenging tasks, such as walkable regions recognition, local path planning, obstacle avoidance, etc. In order to evaluate the system's performance, we employ seven participants equipped with the navigation system to conduct three tasks, including finding specific locations, passing through obstacle regions, and crossing intersections. The results demonstrate that our system can assist users in completing tasks efficiently. In future work, we will focus on researching semi-supervised image segmentation methods to improve the system's performance.

## REFERENCES

[1] L. Chen *et al.*, "Driving with LLMs: Fusing Object-Level Vector Modality for Explainable Autonomous Driving," in *2024 IEEE International Conference on Robotics and Automation (ICRA)*, May 2024, pp. 14093–14100. doi: 10.1109/ICRA57147.2024.10611018.
[2] B. S, S. M, J. J, K. G, and P. Sb, "Face detection in untrained deep neural networks," *Nature communications*, vol. 12, no. 1, Dec. 2021, doi: 10.1038/s41467-021-27606-9.
[3] Y. Sha, "Efficient Facial Landmark Detector by Knowledge Distillation," in *IEEE International Conference on Automatic Face and Gesture Recognition*, Dec. 2021, pp. 1–8. Accessed: Jun. 05, 2024.
[4] Y. Sha *et al.*, "MetDIT: Transforming and Analyzing Clinical Metabolomics Data with Convolutional Neural Networks," *Analytical Chemistry*, 2024.
[5] Y. Su, T. Lan, H. Li, J. Xu, Y. Wang, and D. Cai, "PandaGPT: One Model To Instruction-Follow Them All," May 25, 2023, *arXiv*: arXiv:2305.16355. doi: 10.48550/arXiv.2305.16355.
[6] C. Li *et al.*, "LLaVA-Med: Training a Large Language-and-Vision Assistant for Biomedicine in One Day," *Advances in Neural Information Processing Systems*, vol. 36, pp. 28541–28564, Dec. 2023.
[7] H. Bao, L. Dong, S. Piao, and F. Wei, "BEiT: BERT Pre-Training of Image Transformers," Sep. 03, 2022, *arXiv*: arXiv:2106.08254. doi: 10.48550/arXiv.2106.08254.

[8] T. Huynh-The, Q.-V. Pham, X.-Q. Pham, T. T. Nguyen, Z. Han, and D.-S. Kim, "Artificial intelligence for the metaverse: A survey," *Engineering Applications of Artificial Intelligence*, vol. 117, p. 105581, Jan. 2023, doi: 10.1016/j.engappai.2022.105581.

[9] H. Caesar *et al.*, "nuScenes: A Multimodal Dataset for Autonomous Driving," presented at the Proceedings of the IEEE/CVF Conference on Computer Vision and Pattern Recognition, 2020, pp. 11621–11631. Accessed: Oct. 10, 2024.

[10] P. E. Hart, N. J. Nilsson, and B. Raphael, "A Formal Basis for the Heuristic Determination of Minimum Cost Paths," *IEEE Transactions on Systems Science and Cybernetics*, vol. 4, no. 2, pp. 100–107, Jul. 1968, doi: 10.1109/TSSC.1968.300136.

[11] T. Chen *et al.*, "A Corresponding Region Fusion Framework for Multi-modal Cervical Lesion Detection," *IEEE/ACM Transactions on Computational Biology and Bioinformatics*, 2022.

[12] J. Ho, A. Jain, and P. Abbeel, "Denoising Diffusion Probabilistic Models," in *Advances in Neural Information Processing Systems*, Curran Associates, Inc., 2020, pp. 6840–6851.

[13] Y. Sha *et al.*, "HerbMet: Enhancing metabolomics data analysis for accurate identification of Chinese herbal medicines using deep learning," *Phytochem Anal*, Aug. 2024, doi: 10.1002/pca.3437.

[14] G. Menghani, "Efficient Deep Learning: A Survey on Making Deep Learning Models Smaller, Faster, and Better," *ACM Comput. Surv.*, vol. 55, no. 12, p. 259:1-259:37, Mar. 2023, doi: 10.1145/3578938.

[15] S. Tang *et al.*, "PP-MobileSeg: Explore the Fast and Accurate Semantic Segmentation Model on Mobile Devices," Apr. 11, 2023, *arXiv*: arXiv:2304.05152. doi: 10.48550/arXiv.2304.05152.

[16] S. Mehta, M. Rastegari, L. Shapiro, and H. Hajishirzi, "ESPNetv2: A Light-Weight, Power Efficient, and General Purpose Convolutional Neural Network," presented at the Proceedings of the IEEE/CVF Conference on Computer Vision and Pattern Recognition, 2019, pp. 9190–9200. Accessed: Oct. 20, 2023.

[17] A. Paszke, A. Chaurasia, S. Kim, and E. Culurciello, "ENet: A Deep Neural Network Architecture for Real-Time Semantic Segmentation," Jun. 07, 2016, *arXiv*: arXiv:1606.02147. doi: 10.48550/arXiv.1606.02147.

[18] H. Zhao, X. Qi, X. Shen, J. Shi, and J. Jia, "ICNet for Real-Time Semantic Segmentation on High-Resolution Images," presented at the Proceedings of the European Conference on Computer Vision (ECCV), 2018, pp. 405–420. Accessed: Oct. 20, 2023.

[19] L.-C. Chen, G. Papandreou, I. Kokkinos, K. Murphy, and A. L. Yuille, "DeepLab: Semantic Image Segmentation with Deep Convolutional Nets, Atrous Convolution, and Fully Connected CRFs," *IEEE Transactions on Pattern Analysis and Machine Intelligence*, vol. 40, pp. 834–848, Apr. 2018.

[20] C. Yu, C. Gao, J. Wang, G. Yu, C. Shen, and N. Sang, "BiSeNet V2: Bilateral Network with Guided Aggregation for Real-Time Semantic Segmentation," *Int J Comput Vis*, vol. 129, no. 11, pp. 3051–3068, Nov. 2021, doi: 10.1007/s11263-021-01515-2.

[21] T.-Y. Lin *et al.*, "Microsoft COCO: Common Objects in Context," in *Computer Vision – ECCV 2014*, D. Fleet, T. Pajdla, B. Schiele, and T. Tuytelaars, Eds., Cham: Springer International Publishing, 2014, pp. 740–755. doi: 10.1007/978-3-319-10602-1_48.

[22] M. Hu, S. Wang, B. Li, S. Ning, L. Fan, and X. Gong, "PENet: Towards Precise and Efficient Image Guided Depth Completion," in *2021 IEEE International Conference on Robotics and Automation (ICRA)*, May 2021, pp. 13656–13662. doi: 10.1109/ICRA48506.2021.9561035.

[23] A. Geiger, "Are we ready for autonomous driving? The KITTI vision benchmark suite," in *Proceedings of the 2012 IEEE Conference on Computer Vision and Pattern Recognition (CVPR)*, in CVPR '12. USA: IEEE Computer Society, Jun. 2012, pp. 3354–3361.

[24] B. Zhai *et al.*, "SqueezeWave: Extremely Lightweight Vocoders for On-device Speech Synthesis," Jan. 16, 2020, *arXiv*: arXiv:2001.05685. doi: 10.48550/arXiv.2001.05685.

[25] Y. Sha *et al*. "CerviFusionNet: A Multi-Modal, Hybrid CNN-Transformer-GRU Model for Enhanced Cervical Lesion Multi-classification." iScience, 2024, pp. 111313.

[26] Y. Sha *et al*. "Towards occlusion robust facial landmark detector ," in *Proceedings of the 2012 IEEE Conference on Automatic Face and Gesture Recognition (FG), 2021,* pp. 1-8.

[27] Y. Sha *et al.* "A novel lightweight deep learning fall detection system based on global-local attention and channel feature augmentation," Interdisciplinary Nursing Research. vol. 2, pp. 68-75, 2023.